\documentclass[10pt,letterpaper,twocolumn]{article}
\usepackage[latin9]{inputenc}
\usepackage{array}
\usepackage{url}
\usepackage{multirow}
\usepackage{amsmath}
\usepackage{amssymb}
\usepackage{graphicx}
\usepackage[breaklinks=true,bookmarks=false]{hyperref}

\makeatletter

\pdfpageheight\paperheight
\pdfpagewidth\paperwidth

\providecommand{\tabularnewline}{\\}

\usepackage{cvpr}\usepackage{times}\usepackage{epsfig}

\cvprfinalcopy % *** Uncomment this line for the final submission

\def\cvprPaperID{1321} % *** Enter the CVPR Paper ID here

\ifcvprfinal\fi

\makeatother

\begin{document}
%%%%%%%%% TITLE

\title{Beyond Physical Connections: Tree Models in Human Pose Estimation}

\author{Fang Wang$^{1,2}$ \\
{\tt\small fang.wang@nicta.com.au}
\and
Yi Li$^{2}$ \\
{\tt\small yi.li@nicta.com.au}
\and
$^1$ Nanjing University of Science and Technology, Nanjing, China, 210094 \\
$^2$ National ICT Australia (NICTA), Canberra, Australia, 2601 \\
}

%\author{Fang Wang$^{1,2}$ \hspace*{90pt}  Yi Li$^2$ \\
%\emph{fang.wang@nicta.com.au  \hspace*{40pt} yi.li@nicta.com.au }\\
%\emph{ $^1$ Nanjing University of Science and Technology}\\
%\emph{ $^2$ National ICT Australia (NICTA)}\\

\maketitle

\begin{abstract}
Simple tree models for articulated objects prevails in the last decade.
However, it is also believed that these simple tree models are not
capable of capturing large variations in many scenarios, such as human
pose estimation. This paper attempts to address three questions: 1)
are simple tree models sufficient? more specifically, 2) how to use
tree models effectively in human pose estimation? and 3) how shall
we use combined parts together with single parts efficiently? 

Assuming we have a set of single parts and combined parts, and the
goal is to estimate a joint distribution of their locations. We surprisingly
find that no latent variables are introduced in the Leeds Sport Dataset
(LSP) during learning latent trees
%\cite{Choi:2011:LLT:1953048.2021056}
for deformable model, which aims at approximating the joint distributions
of body part locations using minimal tree structure. This suggests
one can straightforwardly use a mixed representation of single and
combined parts to approximate their joint distribution in a simple
tree model. As such, one only needs to build \textit{Visual Categories}
of the combined parts, and then perform inference on the learned latent
tree. Our method outperformed the state of the art on the LSP, both in
the scenarios when the training images are from the same dataset and
from the PARSE dataset. Experiments on animal images from the VOC challenge
further support our findings.
\end{abstract}
%%%%%%%%% BODY TEXT

\section{Introduction}

Tree models are very efficient in a number of computer vision tasks
such as human pose estimation and other articulated body modeling.
These models prevail because they are simple and exact inference is
available. Also because of these unique advantages, it is not uncommon
to speculate that tree models may not effectively handle computer
vision problems in real world applications. 

As a consequence, latent variables \cite{DBLP:conf/eccv/TianZN12}
and loopy graphical models \cite{wang2011learning} were proposed
in the past few years for human pose estimation as the remedy of the
problems caused by those ``oversimplified'' tree models. All these methods
have their advantages both in theory and practice, and performance
improvements are significant. Particularly, it is believed that loopy
graphical models are necessary when combined parts (or ``poselet'')
are used to handle large variance in appearance. However, no theoretical
analysis shows 1) how to combine parts in component-based learning
for computer vision, and 2) which model is optimal. 

In this paper, we argue that the simple tree model is still a very
powerful representation, and combined parts and single parts can be
used together without sacrificing the benefits brought by exact inference.
We further address how to use it optimally in human pose estimation. 

\begin{figure*}[!t]
\begin{centering}
\begin{tabular}{ccccc}
\includegraphics[height=0.12\textheight]{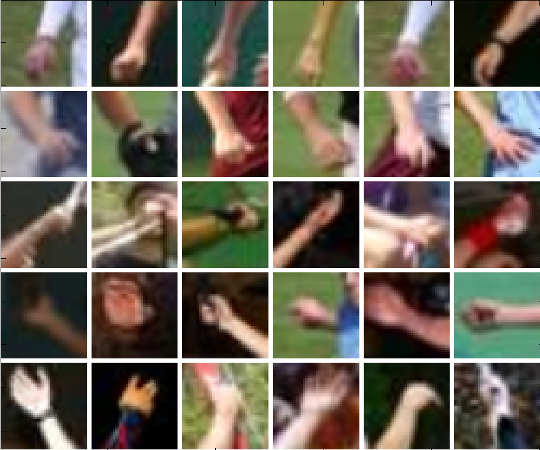} \hspace*{-8pt} & 
\includegraphics[height=0.12\textheight]{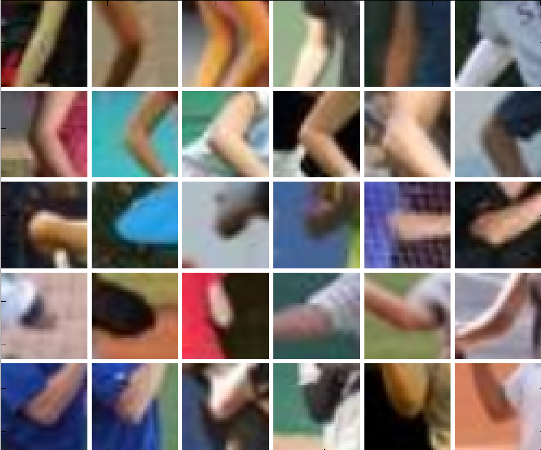} \hspace*{-8pt} & 
\includegraphics[height=0.12\textheight]{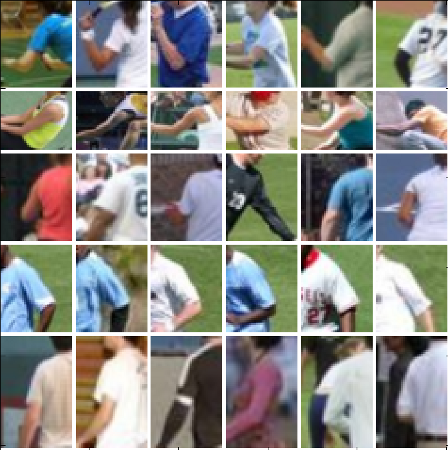} \hspace*{-8pt} &
\includegraphics[height=0.12\textheight]{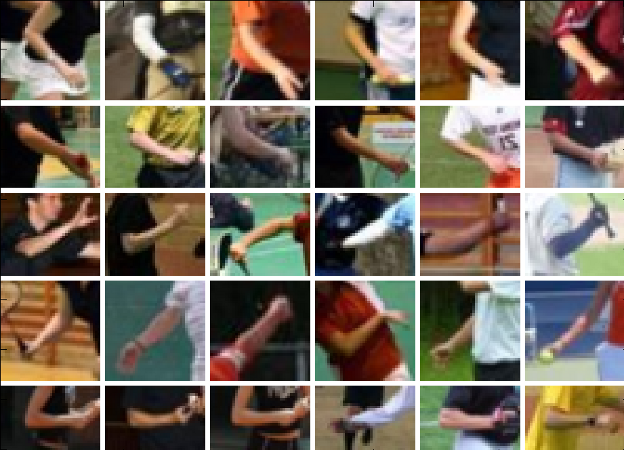} \hspace*{-8pt} & 
\includegraphics[height=0.12\textheight]{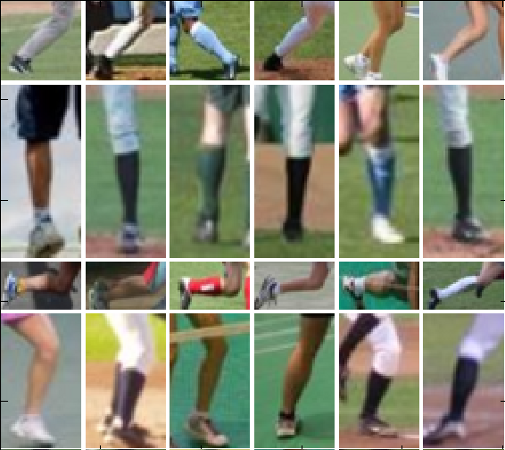}  \tabularnewline
(a) Hand & (b) Elbow & (c) Upper arm & (d) Lower arm  & (e) Lower leg \tabularnewline
\end{tabular}
\par\end{centering}

\caption{Examples of types in the LSP dataset. The types for single parts a) and b)
are defined by their relative positions to their neighbors. For combined
parts c)-e), the types are defined by their visual categories. Each
row represents one type of a part. \label{fig:An-illustration-of}}
\end{figure*}

All the questions about tree models arise when we start to use the
skeleton as the tree structure. While this ``mapping'' is apparent,
no evidence claims it is optimal. Further, this limits our choices
of representation, as well as complicates the graphical model when
combined parts are introduced.

Our goal is to learn a tree model directly from observed variables.
These observations can be body part locations,
as used in many recent pose estimation papers. At the same time, this 
allows us to introduce more variables such as combined parts, as long
as they can be observed and the state space is the same as that of
single body parts.

A significant advantage of our framework is that single joints and
body limbs are considered in the same level in inference. Recent advancements
in learning graphical models enable us to learn \textit{latent trees}
from these observations. The latent tree models suggest that we could
approximate the joint distribution of the observations by a tree model,
and latent variables are introduced only when necessary.

We start our journey by exploring the property of the latent tree
models. It is not surprising that the resulting latent tree has a similar
structure to human skeleton. What really surprising is that
there was \textit{no hidden variable} added when we applied it to the
Leeds Sport dataset (LSP) \cite{Johnson10}. This human pose estimation
dataset is challenging both in the pose variations and its size. 
Therefore, this implies we can directly use a tree model with
mixed types of variables for human pose estimation to approximate
the true distribution.

This exciting claim cannot be achieved without the appearance model.
While the state space of these observations is the same (\textit{i.e.},
spatial location), their appearance models vary. We used ``types'',
a concept in recent papers, to capture the appearance clustering \cite{yang2011articulated,DBLP:conf/eccv/TianZN12}. 

The types were proposed to define the possible configurations between
adjacent nodes in graphical model \cite{yang2011articulated}. We
follow this guideline, but redefine the type being its \textit{Visual
Category} if a node represents a combined part (Fig. \ref{fig:An-illustration-of}).
The concept of visual category is no stranger to computer vision and
pose estimation. Unlike other state of the arts in human pose estimation
that use only a small number of categories, we find that a larger
number of visual categories facilitates pose estimation. This also
makes our framework unlimited to the physical connections \cite{Felzenszwalb:2005:PSO:1024426.1024429},
and flexible to other articulated objects estimation in computer vision.

The inference of our model is very efficient due to the tree structure,
and results suggest that our method outperforms state of the art on
the LSP dataset. We further tested our model in a cross-dataset experiment,
where we used PARSE \cite{ramanan2007learning} dataset for training
and LSP for testing. Our performance does not decrease in this
challenging test. Finally, we verified our hypothesis by testing on a subset
of animal images in the VOC dataset \cite{pascal-voc-2009}.

Our contributions include:
\begin{itemize}
\item We propose to learn tree models for articulated pose estimation problems. 
\item Our method effectively exploits the interactions between combined
parts and single parts.
\item Our method outperforms the state of the art in human and animal pose
estimation.
\end{itemize}

\section{Related work}

\subsection*{Human pose estimation}

Human pose estimation has been formulated as a part based inference
problem. Appearance model and deformable model that describe relations
between parts were proposed in the past decade.

Rich \textbf{appearance models} were adopted extensively in estimating
human poses. Histogram of Oriented Gradient (HOG) \cite{dalal2005histograms}
were frequently used as the features for the appearance model of body
parts. Bourdev \textit{et al.} \cite{bourdev2010detecting} proposed
the idea of poselets as the building blocks for human recognition,
which refers to combined parts that are distinctive in training images.

One successful example of \textbf{deformable models} is the Pictorial
Structure Model (PSM) \cite{Felzenszwalb:2005:PSO:1024426.1024429}.
Pairwise terms in human pose estimation are represented as relative
distance between two parts. Such definition allows efficient distance
transform method to be used in message passing process. Yang \textit{et
al.} \cite{yang2011articulated} proposed a flexible mixtures-of-parts
model for articulated pose estimation. Instead of modeling both location
and orientation of each body part as rigid part, they used the model
that only contains non-oriented parts with co-occurrence
constraints.

It is widely hypothesized that graphical models that go beyond pairwise
links lead to better performance in pose estimation. Loopy graph
model may give more precise results, but it takes more efforts to
solve \cite{wang2011learning}. Several new approaches also use latent
nodes \cite{DBLP:conf/eccv/TianZN12} or hierarchical graph models
\cite{sun2011articulated}. 

In this paper, we examine the above concepts, and suggest that these
important components in articulated body detection and pose estimation
can be integrated in an efficient framework. Therefore, we exploit
a newly developed technique in learning latent tree models.

\subsection*{Latent tree models}

The latent tree models \cite{Choi:2011:LLT:1953048.2021056} aim at
finding tree approximations of joint distribution of observable variables. Using trees
to approximate joint distribution has been dated to the early days
of machine learning. In Chow-Liu tree \cite{Chow1968} all nodes in
the latent tree must be observable. Recently, Choi \textit{et al.} \cite{Choi:2011:LLT:1953048.2021056}
proposed two algorithms on learning latent trees.
Their methods automatically build tree structures from observations,
using information distances as the guideline of merging nodes and
introducing latent variables. 

This theoretical approach is very useful for human pose estimation,
because we can learn a structure directly from our observations without
making many assumptions of the physical constraints, while the performance
is still guaranteed in terms of approximating the joint distribution.

\section{Latent tree models for pose estimation}

First, we provide a brief introduction to the latent tree models,
and show the results on modeling the body joints in the LSP using
latent tree models. We then present the learning of visual category
for combined parts, which is necessary in our model for reducing complexity.
Finally, we present our tree based inference model.

\begin{figure}[!tp]
\begin{centering}
\includegraphics[width=0.95\columnwidth]{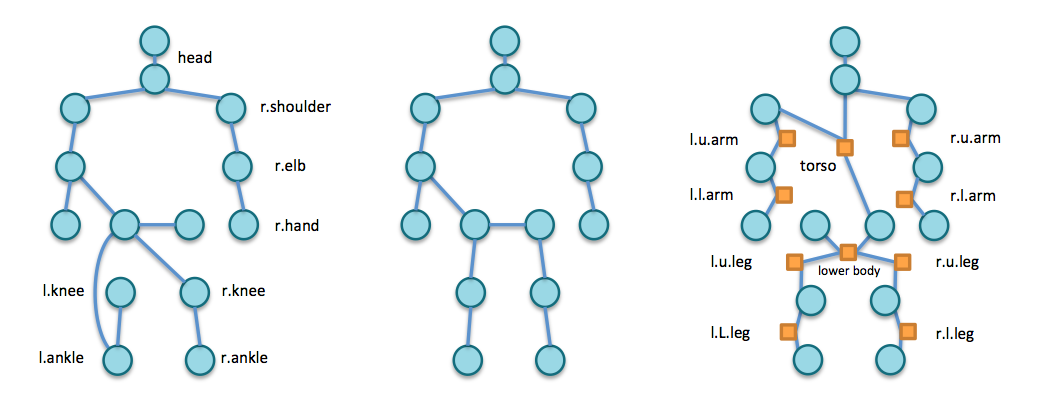}
\par\end{centering}

\caption{Latent tree models learned from the LSP dataset. From left to right,
the results for CLGrouping tree,
CL-Neighbor Joining \cite{Choi:2011:LLT:1953048.2021056} using single parts, respectively,
and CLGrouping on single and combined parts together. The circles
denote single parts, and the squares denote combined parts. \label{fig:Latent-tree-results}}
\end{figure}

\subsection{Brief introduction to latent tree models}

The goal of latent tree model is to recover a tree-structured graphical model that best approximates the distributions of a set of observations. 
Recursive Grouping and CLGrouping  were proposed in \cite{Choi:2011:LLT:1953048.2021056} 
to create latent tree models without any redundant hidden nodes. 
The authors demonstrated their methods on real-world datasets by modeling
the dependency structure of monthly stock returns in the S\&P index
and of the words in the 20 newsgroups dataset.
Please note that in the resulting models, the observed variables can be both leaf nodes
and non-leaf nodes. 

We briefly describe recursive grouping in this section due to its
simplicity.
CLGrouping is its extension that can build up latent tree structures for large diameter graphs
more efficiently with a pre-processing step. 

In this grouping method, the latent tree were built recursively
by identifying sibling groups using information distances. Given two
observed random variables $X_{i}$ and $X_{j}$, the correlation coefficient
is defined as 
\begin{equation}
\rho_{ij}=\frac{\text{Cov}(X_{i},X_{j})}{\sqrt{\text{Var}(X_{i})\text{Var}(X_{j})}}
\end{equation}
and the information distance is defined as
\begin{equation}
d_{ij}=-\text{log}(\rho_{ij})
\end{equation}

%In the discrete case, this distance is defined on the entropy instead.
Then, the recursive grouping method build up the latent tree by testing relationships 
among each triplet $i,j,k \in V$. Define $\Phi_{ijk}\triangleq d_{jk}-d_{ik}$, take one
of the two actions below:
\begin{itemize}
\item If $\Phi_{ijk}=d_{ij},$ $j$ is set to be the parent of $i$.
\item If $-d_{ij}\leq\Phi_{ijk}=\Phi_{ijk'}\leq d_{ik}$ for all $k$ and
$k'\in V\backslash\{i,j\}$, add a hidden node as the parent of $i$
and $j$.
\end{itemize}
In this way, a latent tree is recursively built. 
Please refer to \cite{Choi:2011:LLT:1953048.2021056} for details.

\subsection{Latent trees for human pose}

Our goals is to use single parts and combined parts in the inference
model. Given image $I$, we define $P$ parts as $p_{i}=(loc_{i},t_{i}),i\in[1,...,P]$,
where $loc_{i}$ is the part location in images, $t_{i}$ is either visual
category label for combined parts (Sec \ref{sub:Visual-category-for}),
or represents different morphologies of parts for single
parts as suggested in \cite{yang2011articulated}. Two possible part
combinations in our case are:
\begin{itemize}
\item \textbf{Connected parts}: A combined part may have physical connection
in human body. This is a natural choice for many problems, because
connected parts (\textit{e.g.}, upper arm and lower arm) may have
higher correlations in general.
\item \textbf{Physically separated parts}: The combined parts can be used
for encoding semantic relations among single parts. This can be totally
data driven. For instance, in many applications arm poses are symmetric.
Therefore, one may combine these two physically separated parts as
one element.
\end{itemize}
In our following experiment, we defined 14 single parts and 10 combined
parts (Fig. \ref{fig:Latent-tree-results}). We used their spatial
correlation in the image space as the mutual information, and Fig.
\ref{fig:Latent-tree-results} shows the results by three different
algorithms. We tested two scenarios in our experiment:
\begin{itemize}
\item \textbf{Single parts only}: In this experiment, we used only single
parts for the latent tree models. Fig. \ref{fig:Latent-tree-results}
shows two results using CLGrouping tree and CL-Neighbor Joining \cite{Choi:2011:LLT:1953048.2021056}. It is not very surprising that
the structure is similar to human body, but please note that there
is no latent node introduced by CLGrouping method in such a complicated
and challenging dataset. Because no redundant latent nodes is used
in latent tree model, this means the joint distributions of all body
joints can be approximated by a simple tree structure.
\item \textbf{Single+combined parts}: Due to the limitation of appearance
model, it is more effective to use combined parts in appearance model
for detecting parts. Therefore, introducing combined parts is a solution
in many algorithms. We used both single and combined parts in the
latent tree models (Fig. \ref{fig:Latent-tree-results}). Again, the
output has no latent variables. This means we can approximate the
joint distribution by combined parts and single parts in a tree structure.
\end{itemize}
This finding makes our latent tree model different from \cite{DBLP:conf/eccv/TianZN12},
because all our nodes are observable. Also, this is different from
\cite{wang2011learning}, because our structure is a tree.

\begin{figure*}[!t]
\begin{centering}
\begin{tabular}{c}
 \includegraphics[width=0.95\textwidth]{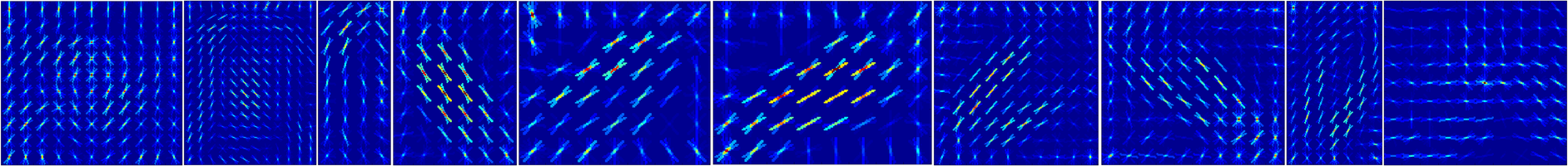}\tabularnewline
 \includegraphics[width=0.95\textwidth]{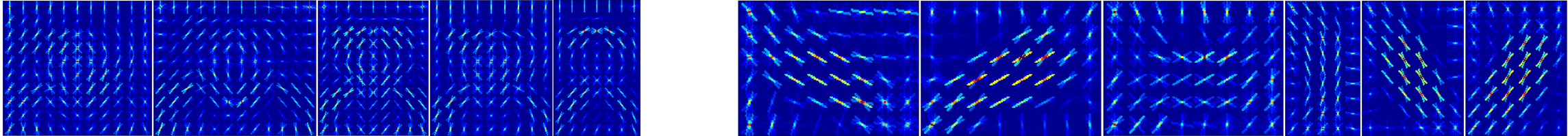}\tabularnewline
\end{tabular}.
\par\end{centering}

\caption{Examples of visual category. Top: the HOG template
for different combined parts (from left to right, head, torso, left
upper leg, left lower leg, right upper leg, right lower leg, left
upper arm, left lower arm, right upper arm, and right lower arm, respectively),
Bottom: HOG template for visual categories for head and left lower leg,
respectively. \label{fig:LSP-1}}
\end{figure*}

\subsection{Learning visual categories of combined parts\label{sub:Visual-category-for}}

Combined parts are more discriminative than single parts. However,
human pose could be very flexible. Even given only two parts, the
limb configurations still span in a very large state space. In order
to represent this large variation of part combination, we use Visual
Category to represent the combined parts. 

We learned visual categories of each combined part directly from image
space. Instead of using semantics or empirical heuristics, we use
appearance-based clustering for generating the categories. This is different from the DPM models
(\cite{felzenszwalb2010object}), where each category comprises
of heuristic rules such as left-right flipping and aspect-ratio. 
Also, this is different from those techniques that cluster combined part
according to relative positions of labelled points (\textit{e.g.},
\cite{wang2011learning}), or learning prior for different configurations \cite{Johnson10}. 
Our strategy makes the body part detectors
more effective. As a result, each combined part has multiple categories.
This reduces its state space but also maintains its representation
power.

For each part, we build a latent SVM (\cite{DBLP:journals/corr/abs-1206-3714})
model for learning visual categories. 
We run a simple $k$-means algorithm on geometric configuration to find mean patch sizes of the same part and crop the image patches.
These mean sizes may be used to normalize the filters learned by the following procedure.

Given $N$ instances of a combined part, we learn $K$ categories of this part, and generate
the label set $T={t_{1},t_{2},\cdots,t_{N}},t_{i}\in[1,K]$. Our objective
function for each category is as follows 
\begin{equation}
\begin{split}\arg\min_{w}\frac{1}{2}\sum_{k=1}^{K}\parallel w_{k}\parallel^{2}+C\sum_{i=1}^{N}\epsilon_{i},\\
y_{i}w_{t_{i}}\phi(x_{i})\geq1-\epsilon_{i},\epsilon_{i}\geq0,\\
t_{i}=\arg\max_{k}w_{k}\phi(x_{i})
\end{split}
\end{equation}

where $\phi(x_{i})$ is the feature map of an image patch $x_{i}$, $y_i$
is the training labels, and $w_k$ is the learned weights of the feature
map for each combined part. 

The visual categories of combined parts characterize the appearance
models in a way that they can be regarded as ``templates''. Therefore,
these filters are also called HOG template, when $\phi(x)$ is HOG
filter. We show the results of HOG filters for different parts as
well as different visual categories for two parts in Fig. \ref{fig:LSP-1}.

\subsection{Our model\label{sub:Our-model}}
Given a training set, we manually define the parts of interest, and
learn a latent tree model for these parts. The following notations
are consistent with those in \cite{yang2011articulated}, while our types for combined parts have different
meanings.

\begin{figure*}[!t]
\begin{centering}
\begin{tabular}{c}
\includegraphics[width=0.97\textwidth]{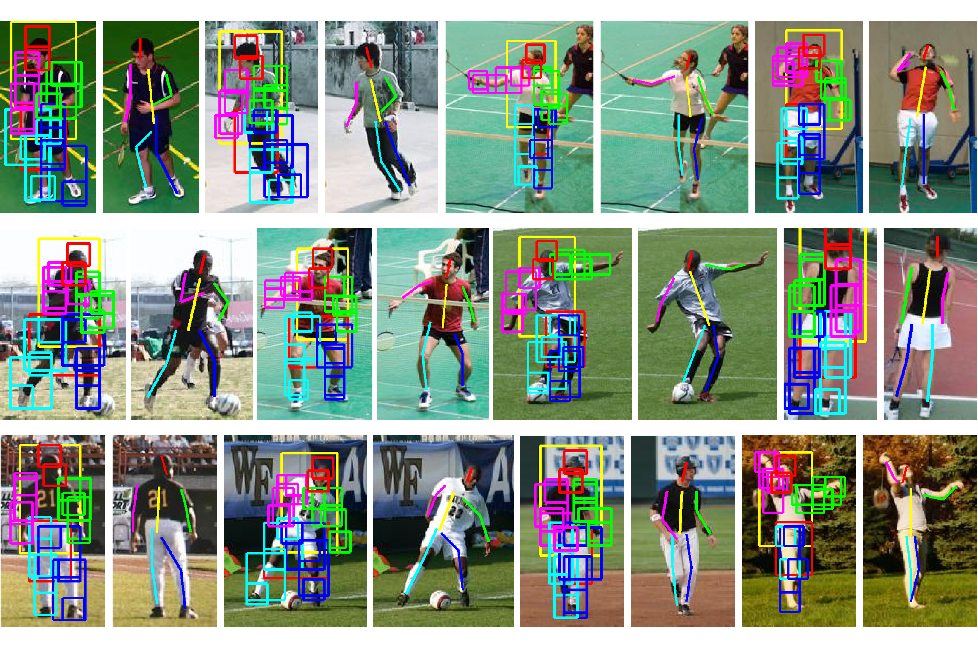}\tabularnewline
\end{tabular}
\par\end{centering}

\caption{Our results on the LSP dataset. We show the bounding box of the part
detection (left) as well as the fitted skeletons (right). Color: Yellow:
torso; Blue: left leg; Cyan: right leg; Green: left arm; Magenta: right arm; Red: head. \label{fig:LSP}}
\end{figure*}

\paragraph*{Appearance term}
Following Felzenszwalb \textit{et al.} \cite{felzenszwalb2010object},
we represent each visual category as a HOG template. For each location
$loc_{i}$ in image $I$, the appearance score of local
patch can be written as 
\begin{equation}
S(I,p_{i})=\omega_{i}^{t_{i}}\phi(I,loc_{i}),
\end{equation}
where $\omega_{i}^{t_{i}}$ is HOG template for category $t_{i}$,
$\phi(\cdot)$ is the HOG feature.

\paragraph*{Deformable term}
Pairwise term between connected part $p_{i},p_{j}$ is defined as
\begin{equation}
S(I,p_{i},p_{j})=\omega_{ij}^{t_{i}t_{j}}\psi(p_{i},p_{j}),
\end{equation}

where $\psi(p_{i},p_{j})=[dx,dy,dx^{2},dy^{2}]$, a space invariant
definition of deformable component. This term can be computed effectively
by distance transform in inference.

\paragraph{Compatibility term}
Compatibility, or co-occurrence, is defined as 
\begin{equation}
S(t)=\sum b_{i}^{t_{i}}+\sum b_{ij}^{t_{i}t_{j}}
\end{equation}

This term denotes whether two types are compatible in the training
set. Once learned, this term greatly reduces the search space in inference.
Note that this term is decomposable to unary terms and pairwise terms.

\paragraph{Objective function}
Our objective function is as follows
\begin{equation}
p=\arg\max_p S(t)+\sum_{i}S(I,p_{i})+\sum_{i,j}S(I,p_{i},p_{j})
\end{equation}

Since our model is a tree, this is a textbook example of exact inference,
and standard message passing algorithm is applicable.

\paragraph{Learning model parameters}

Denote the model parameter as $\beta$, which consists of HOG filters
for single parts and deformable models.
The learning amounts to the quadratic optimization as follows
\begin{equation}
\begin{split}\arg\min_{\beta,\xi_{i}\geq0}\frac{1}{2}\Vert\beta\Vert_2 +C\sum_{n=1}^{N}\xi_{i},\\
\forall n\in\text{pos}\quad\langle\beta,\Phi(I_{n},p)\rangle\geq1-\xi_{n},\\
\forall n\in\text{neg}\quad\langle\beta,\Phi(I_{n},p)\rangle\leq-1+\xi_{n}.
\end{split}
\end{equation}
%In this learning process, the positive
%examples must be greater than 1, while negative examples being less than -1. 
where $I_{n}$ as the image, $\Phi(I_{n},p)$ as the concatenated features of given instance $p$. 
This is a standard quadratic programming procedure, and can
be solved effectively. 

\begin{figure*}[ht]
\begin{centering}
\begin{tabular}{c}
\includegraphics[width=0.95\textwidth]{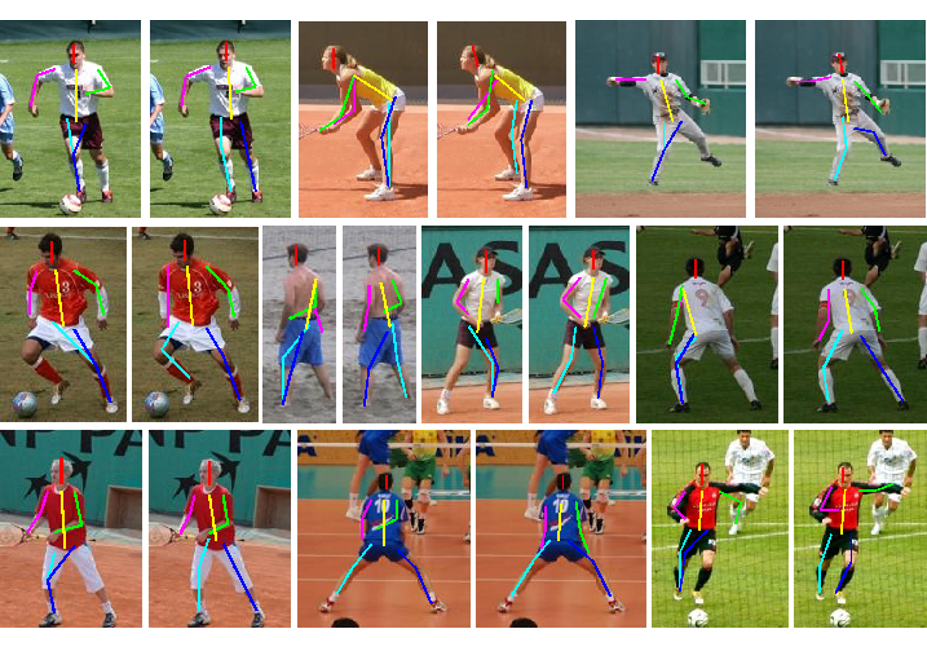}\tabularnewline
\end{tabular}
\par\end{centering}

\caption{Side by side comparison of \cite{yang2011articulated} (left) and our pose estimation results (right) in the Leeds Sport dataset (LSP).
The color notations are consistent with Fig. \ref{fig:LSP}\label{fig:Comparison}}
\label{fig:devacmp} 
\end{figure*}

\section{Experiment results}
We present three experiments in this section. First, we evaluate our
performance on the LSP dataset. Then, we show that our method
performs well in a cross dataset experiment, which suggests our model
does not overfit. Finally, we test our method on a subset of the animal
images in the VOC challenge.

Our single parts are the same as those 14 joints used in \cite{yang2011articulated}.
Our combined parts are defined as the limbs in \cite{DBLP:conf/eccv/SappTT10}.
In all experiments, we firstly extract bounding boxes for all parts
in the training sets. For each combined part, we extract HOG features
on grid image with $4\times4$ pixels from image patches, and learn
visual categories using latent SVM. Then we train the whole
human pose model the procedure defined in Sec. \ref{sub:Our-model}.

In all experiments, we use the negative set of INRIA person dataset \cite{dalal2005histograms} as our negative set,
which have 1218 images of various scenes.
The evaluation criterion is the same with \cite{DBLP:conf/cvpr/FerrariMZ08}
for performance comparison. A part is correctly detected if both its
endpoints are within $50\%$ of the length of corresponding ground
truth segments.

The computational complexity is in the same order of magnitude of that of \cite{yang2011articulated}.
The running time for testing is approximately 2s per image on a Linux 64 bit OS using Core i7 2.2G CPU, with non-optimized Matlab code.

\subsection{LSP dataset}
Leeds Sport Dataset (LSP) \cite{Johnson10} contains 2000 images collected
from various human activities. This dataset has a large variation
of pose changes. Humans in each image were cropped and scaled to 100
to 150 pixels in height. The dataset was split evenly into training
set and testing set, each of which has 1000 images. 

\subsubsection*{Experiment setting}
We used 8-15 visual categories for combined parts.
The images and labels in the training set were manually flipped to increase the variations.
\begin{table*}[!t]
\begin{centering}
\begin{tabular}{|c|c|c|c|c|c|c|c|c|}
\hline 
Exp.  & Method  & Torso  & Head  & U.Leg  & L.Leg  & U.Arm  & L.Arm  & Total \tabularnewline
\hline 
\hline 
%\multirow{7}{*}{LSP} & Yang \& Ramanan \cite{yang2011articulated} & 77.3  & 61.9  & 62.5  & 54.4  & 56.3  & \textbf{39.0 } & 56.4 \tabularnewline
\multirow{7}{*}{LSP} & Yang \& Ramanan \cite{yang2011articulated} & 92.6  & 87.4  & 66.4  & 57.7  & 50.0  & 30.4 & 58.9 \tabularnewline
\cline{2-9} 
 & Tian \textit{et al.} (First 200) \cite{DBLP:conf/eccv/TianZN12} & 93.7  & 86.5  & 68.0  & 57.8  & 49.0  & 29.2  & 58.8 \tabularnewline
\cline{2-9} 
 & Tian \textit{et al.} (5 models) & \textbf{95.8 } & \textbf{87.8 } & 69.9  & 60.0  & \textbf{51.9}  & 32.8  & 61.3 \tabularnewline
\cline{2-9} 
 & Johnson \& Everingham \cite{Johnson10}  & 78.1  & 62.9  & 65.8  & 58.8  & 47.4  & \textbf{32.9}  & 55.1 \tabularnewline
%\cline{2-9} 
% & Johnson \& Everingham 2\cite{DBLP:conf/cvpr/JohnsonE11} & 88.1 & 74.6 & \textbf{74.5} & 66.5 & 53.7 & 37.5 & 62.7\tabularnewline
\cline{2-9} 
 & Ours (First 200)  & 88.4  & 80.8  & 69.1  & 60.0  & 50.5  & 29.2  & 59.0 \tabularnewline
\cline{2-9} 
 & Ours & 91.9  & 86.0  & \textbf{74.0} & \textbf{69.8} & 48.9 & 32.2  & \textbf{62.8}\tabularnewline
\hline 
\hline 
\multirow{2}{*}{Cross dataset } 
 & Yang \& Ramanan \cite{yang2011articulated}  & 78.8  & 70.0  & 66.0  & 61.1  & \textbf{61.0}  & \textbf{37.4}  & 60.0 \tabularnewline
\cline{2-9} 
\cline{2-9} 
 & Ours  & \textbf{88.3}  & \textbf{78.7}  & \textbf{75.2}  & \textbf{71.8}  & 60.0  & 35.9  & \textbf{65.3} \tabularnewline
\hline 
\end{tabular}
\par\end{centering}

\caption{Performance on the LSP dataset. The first 6 rows show the performance
of four algorithms when the training and the testing are both from
the LSP dataset. The last two rows show the comparison between our
method and \cite{yang2011articulated} in a cross dataset experiment.\label{tab:Comparison-results.}}
\end{table*}

\subsubsection*{Results}
\begin{figure}[t]
\begin{centering}
\begin{tabular}{c}
\includegraphics[width=0.9\columnwidth]{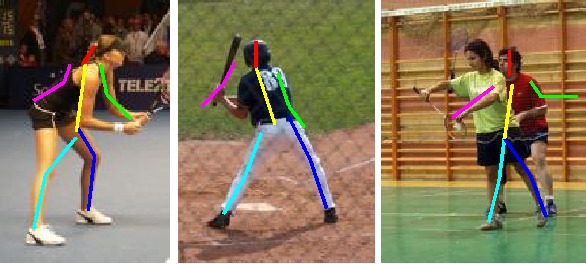}\tabularnewline
\end{tabular}
\par\end{centering}

\caption{Failure examples of our method in the LSP dataset. \label{fig:Failure-examples-in}}
\end{figure}

Fig. \ref{fig:LSP} displays our results on the LSP dataset. The left
image in each pair denotes the detection results for single and combine
parts, and the right is the fitted skeleton. Fig. \ref{fig:Comparison}
shows the comparison between  \cite{yang2011articulated}
(left) and our method (right). We also show some failure examples in Fig. \ref{fig:Failure-examples-in}.

We compared the detection accuracy of our method with Yang \& Ramanan
\cite{ramanan2007learning}, Johnson \& Everingham  \cite{Johnson10},
and Tian \textit{et al.} \cite{DBLP:conf/eccv/TianZN12} respectively.
Table \ref{tab:Comparison-results.} summarizes the evaluation results
and highlights the highest scores.

Compared to Yang \& Ramanan \cite{ramanan2007learning}, our results are better.
Our detection accuracies on Upper Leg ($74.0\%$), Lower Leg ($69.8\%$), and Total ($62.8\%$) are consistently higher. 
Our performance is also superior to Johnson \& Everingham \cite{Johnson10}.
We are aware of their later method \cite{DBLP:conf/cvpr/JohnsonE11}, which achieved $62.7\%$ in total accuracy, but they trained their model on 11000 samples and relabeled the training set during optimization.

Recently, Tian \textit{et al.} \cite{DBLP:conf/eccv/TianZN12} suggested to partition all the 1000 training
images into 5 disjoint training sets, and the detection accuracy is defined as the maximal score of 5 models.
This approach is practical, because the evaluation method in \cite{DBLP:conf/cvpr/FerrariMZ08} prompts the algorithms that report more candidate detections.
With this evaluation bias, our method still marginally outperforms theirs.
We further report our detection results using the first 200 training images ($6^{th}$ row) for fair comparison.

This experiment suggests our method outperformed the state of the art. This is
possibly because we effectively exploit the connections between single
and combined parts, as well as the benefit from exact inference.

\subsection{Cross dataset validation}
\subsubsection*{Experiment setting}
We further investigate the generalization power of our model by cross
dataset validation. One may speculate that our combined parts may
be overfitted to a dataset, because they captures the distinctive
features as HOG templates during visual category learning. This experiment
suggests that our model is able to generalize to different dataset.

We trained our model on all the 305 images in the PARSE dataset \cite{ramanan2007learning}, and
then used the models to estimate human pose on the LSP dataset.
We manually relabeled the LSP dataset for the purpose of testing, because its definition of left/right is based on human coordinate while the PARSE dataset is based on image coordinate.
We used the same setting for Yang \& Ramanan \cite{yang2011articulated}.

\begin{figure*}[ht]
\begin{centering}
\begin{tabular}{c}
\includegraphics[width=0.97\textwidth]{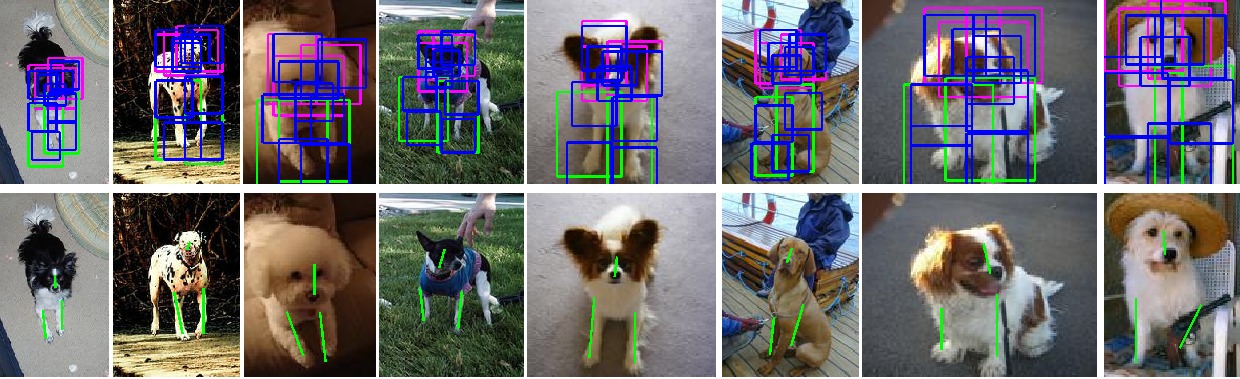}\tabularnewline
\end{tabular}
\par\end{centering}

\caption{Examples of our method on the dog dataset from Pascal 2009. Top images:
detection results for head (magenta), foreleg (green and blue). Bottom
images: fitted skeleton after inference.\label{fig:Comparison-1-1}}

\label{fig:devacmp-1-1} 
\end{figure*}

\subsubsection*{Results}
The last two rows in Table \ref{tab:Comparison-results.} shows the
detection accuracies of both algorithms. Compared to the $7^{th}$
row and $9^{th}$row, the performance of our model does not degrade, and surprisingly
produce higher accuracy in some parts. 
As a result, our method empirically outperformed \cite{yang2011articulated}
in four out of six joints, and the Total accuracy is $5\%$ higher.
Therefore, we conclude that the combination of mixed types facilitate
inference process, and avoid overfitting in datasets.

\subsection{Pascal VOC dog dataset}
We tested our method on dog images from the PASCAL 2009 dataset.
Bourdev \textit{et al.} \cite{bourdev2010detecting} annotated these images with up to 26 keypoints, but not all images
have all annotations. In our experiment, we select 280 images that
have at least the following 9 keypoints, namely, ``nose'', left
and right ``eye'', ``ear base'', ``front elbow'', and ``front
paw'', respectively. 

\subsubsection*{Experiment setting}
In this experiment we used 3 combined parts, ``head'', ``left leg'', and ``right leg''.
We used 6 visual categories for each combined part.
100 images were used for training and the remaining for testing. We compared our method
to \cite{yang2011articulated} in this experiment.
In the Yang \& Ramanan \cite{yang2011articulated}, we used the 9 keypoints in a natural skeleton structure  to build the tree model.

\begin{table}
\begin{centering}
\begin{tabular}{|c|c|c|c|c|c|}
\hline 
Method  & Head  & L.F.Leg  & R.F.Leg  & Legs & Total \tabularnewline
\hline 
\hline 
\cite{yang2011articulated} & \textbf{56.1}  & 52.8  & 58.3  & 55.6 & 55.7 \tabularnewline
\hline 
Ours  & 52.8  & \textbf{60.6}  & \textbf{63.3}  & \textbf{62.0} & \textbf{58.9} \tabularnewline
\hline 
\end{tabular}
\par\end{centering}

\caption{Results on the dog images, a subset of the VOC challenge.\label{tab:Results-on-the}}
\end{table}

\subsubsection*{Results}
Fig. \ref{fig:Comparison-1-1} shows some typical results in this
experiment. The images on top is the detection results and the ones
in bottom contain the fitted skeletons for dogs. This subset of the
VOC dataset is very challenging, because the large variations in dog
poses and camera viewpoints. Therefore, both \cite{yang2011articulated}
and our algorithm have low accuracies (Table. \ref{tab:Results-on-the}). 

Although our performance is worse in the ``head'', our detection
on the ``left fore leg'' and the ``right fore leg'' are higher ($60.6\%$ and $63.3\%$, respectively). 
An average accuracy for legs is approximately $6\%$ higher than that of \cite{yang2011articulated},
and the total accuracy is $3\%$ higher. This experiment demonstrates
that our method serves as a very good tool for modeling parts
in other articulated objects such as animals.

\section{Conclusion}
This paper addressed three questions in human pose estimation using
deformable models. Latent tree models are learned to approximate the
joint distributions of body part locations, and single and combined
parts are used together for effective inference. 
Empirical results suggest that our approach outperforms the state
of the art in human pose and animal pose estimation.

{\small 
\bibliographystyle{IEEEbib}
\bibliography{cvpr13-wang-pose}
 }
\end{document}